\pdfoutput=1
\documentclass[letterpaper]{article} 
\usepackage[preprint]{aaai2027}  
\usepackage[hyphens]{url}  
\usepackage{graphicx} 
\usepackage{natbib}  
\usepackage{caption} 
\usepackage{amsmath}
\usepackage{amssymb}
\usepackage{booktabs}
\newcommand{\eg}{\textit{e.g.}}
\newcommand{\methodname}{SEAM}

\title{SEAM: Smooth Execution of Action-Chunked Motion \\for Vision-Language-Action Policies}
\author{
  Dijia Zhan$^*$, Xuemiao Xu$^\dagger$, Jinyi Li$^*$, Jie Tang$^\dagger$
}
\affiliations{
  South China University of Technology\\
  $^*$Equal contribution \quad $^\dagger$Corresponding authors
}

\begin{document}

\maketitle

\begin{abstract}
Vision-Language-Action (VLA) policies that execute fixed-length action chunks can exhibit \textit{multimodal bifurcation}: a cross-chunk inconsistency in which adjacent chunks generated from independent Gaussian latents can converge to incompatible trajectory modes, producing abrupt discontinuities at chunk boundaries. Existing remedies either require backpropagation through the policy at each denoising step, rely on rejection sampling, or require retraining, each trading computational cost or task reliability for smoother transitions. We propose \textbf{SEAM} (\textbf{S}mooth \textbf{E}xecution of \textbf{A}ction-chunked \textbf{M}otion), a training-free inference-time method for flow matching VLAs. SEAM exploits a simple synchronous-execution insight: after the robot consumes the executed prefix, the previous chunk's unexecuted tail is already available as an analytic consistency reference. Its core mechanism, \textit{Velocity-guided Loss Steering} (VLS), derives a time-dependent target from this tail and applies a closed-form correction after each Euler step without backpropagating through the policy network. On LIBERO-10 with $\pi_{0.5}$, SEAM reduces boundary jerk by 28\%, reduces chunk transition discontinuity by 27\%, preserves baseline-level task success, and keeps denoising-loop cost near the unguided baseline.
\end{abstract}

\section{Introduction}

\begin{figure*}[t]
\centering
\includegraphics[width=\textwidth]{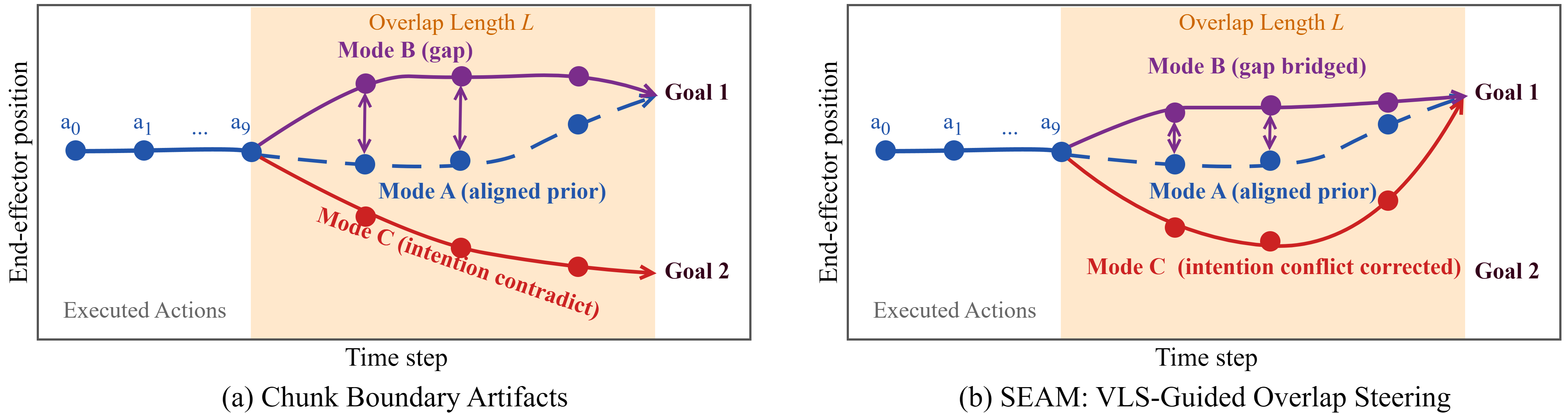}
\caption{%
SEAM overview.
(a) Independent chunk sampling can produce two boundary artifacts: an intention conflict (\textit{Mode C}), where the next chunk reverses motion, and a position gap (\textit{Mode B}), where the next chunk preserves direction but starts from an offset.
Goals 1 and 2 denote two feasible local continuations.
(b) SEAM performs VLS-guided overlap steering by nudging the guided overlap window toward the aligned prior, the unexecuted tail of the previous chunk, bridging the position gap and correcting the intention conflict without backpropagating through the policy.
}
\label{fig:overview}
\end{figure*}

Vision-Language-Action (VLA) policies have become central to language-conditioned robotic manipulation, from RT-1 and RT-2 to OpenVLA, $\pi_0$, and $\pi_{0.5}$~\citep{brohan2022rt1, brohan2023rt2, kim2024openvla, black2024pi0, pi05_2025}. Given visual observations and a natural-language instruction, these policies generate robot action trajectories that must remain task-directed and physically smooth. Recent flow matching VLAs such as $\pi_{0.5}$~\citep{pi05_2025} model this generation process as an ODE that transports Gaussian noise into an action trajectory.

A key design choice in these policies is \textit{action chunking}~\citep{zhao2023act, chi2023diffusion}: the model predicts $H$ future actions at once, executes only the first $K \ll H$, and then predicts the next chunk. Chunking improves local temporal coherence because each prediction contains a short motion plan, but because only $K$ of the $H$ predicted actions are executed before re-prediction, consecutive chunks share an overlap region of $H{-}K$ steps. Ideally, the beginning of the new chunk should agree with the unexecuted tail of the previous one. In practice, however, the two chunks are generated from independent noise samples $\mathbf{z}_n, \mathbf{z}_{n+1} \sim \mathcal{N}(\mathbf{0}, \mathbf{I})$, so their overlap predictions can disagree even when the observations are nearby in time.

This disagreement can manifest as \textit{multimodal bifurcation} (Figure~\ref{fig:overview}a), a cross-chunk inconsistency in which adjacent chunks choose incompatible trajectory modes. Contact-rich manipulation often admits multiple valid local strategies~\citep{shafiullah2022bet}, and a flow matching policy represents these alternatives by mapping different noise samples to different trajectory modes. Adjacent chunks can therefore select incompatible modes: one chunk may continue an approach from one side of an object, while the next begins a different approach, creating a sharp reversal exactly at the chunk transition~\citep{chunkboundary2025, black2025rtc}. This artifact is not merely cosmetic; it increases jerk and discontinuity at the chunk boundary, which can introduce large motion discontinuities and reduce execution reliability.

Existing cross-chunk remedies occupy different points on a cost--smoothness spectrum. They share a common objective: reducing chunk-boundary artifacts by encouraging the new chunk to agree with the unexecuted tail of the previous chunk, but they instantiate this objective through different mechanisms and incur different costs. Temporal ensembling~\citep{zhao2023act} averages overlapping predictions, which can create invalid intermediate actions when adjacent chunks choose incompatible modes. RTC~\citep{black2025rtc} treats continuation as conditional inpainting and applies $\Pi$GDM gradients inside the denoising ODE; this strongly improves consistency, but the required policy-network backpropagation at each ODE step substantially increases denoising latency. BID~\citep{liu2024bid} searches for consistent continuations by rejection sampling, increasing inference cost, while Legato~\citep{liu2026legato} learns continuation-aware dynamics during training and therefore requires retraining the action model. Although these methods demonstrate that cross-chunk consistency can reduce chunk-boundary artifacts, they do not provide a computationally lightweight, training-free analytical mechanism for imposing such consistency.

These limitations suggest a more specific gap in the common synchronous chunked-execution setting: when the robot consumes $K$ actions before requesting the next chunk, the full unexecuted tail of the previous prediction is already available, yet existing remedies either average across incompatible modes, require costly gradient-based inpainting, rely on rejection sampling, or modify the trained policy.

We propose \textbf{SEAM} (\textbf{S}mooth \textbf{E}xecution of \textbf{A}ction-chunked \textbf{M}otion), a training-free inference-time method whose core mechanism is \textbf{Velocity-guided Loss Steering (VLS)}. \methodname{} exploits this overlooked structure by using the previous prediction's $L{=}H{-}K$ tail as an analytic cross-chunk consistency reference. VLS constructs a time-interpolated consistency target from this reference and applies a closed-form correction after each Euler step of the flow matching ODE without backpropagating through the policy network. On LIBERO-10 with $\pi_{0.5}$, the baseline boundary jerk is more than twice its interior jerk (0.195 vs.\ 0.094), indicating that motion irregularity is concentrated at chunk boundaries rather than uniformly distributed across the trajectory. SEAM reduces boundary jerk by 28\% and chunk transition discontinuity by 27\% at $1.01\times$ denoising-loop cost, while preserving baseline-level task success (95.7\% vs.\ 94.8\%). RTC reaches lower boundary jerk and chunk discontinuity with 95.1\% success, but its policy-network gradients raise denoising-loop cost to $1.22\times$ the baseline; ACT-TE reaches lower motion metrics at substantially lower task success. These comparisons highlight SEAM's intended operating regime: improving cross-chunk motion quality while keeping denoising-loop overhead small and preserving task-directed policy predictions.

In summary, our key contributions are as follows:
\begin{itemize}
    \item We cast cross-chunk consistency in synchronous chunked execution as an analytic overlap-alignment objective and propose \methodname{}, a training-free inference-time method to mitigate this artifact.
    \item We design VLS, the core mechanism of \methodname{}, as a closed-form correction that steers the denoising ODE toward an aligned prior without backpropagating through the policy network, reducing cross-chunk inconsistency at execution boundaries.
    \item We validate SEAM on LIBERO-10, showing a 28\% reduction in boundary jerk and a 27\% reduction in chunk discontinuity while preserving baseline-level task success at near-baseline denoising-loop cost.
\end{itemize}


\begin{figure*}[t]
\centering
\includegraphics[width=\textwidth]{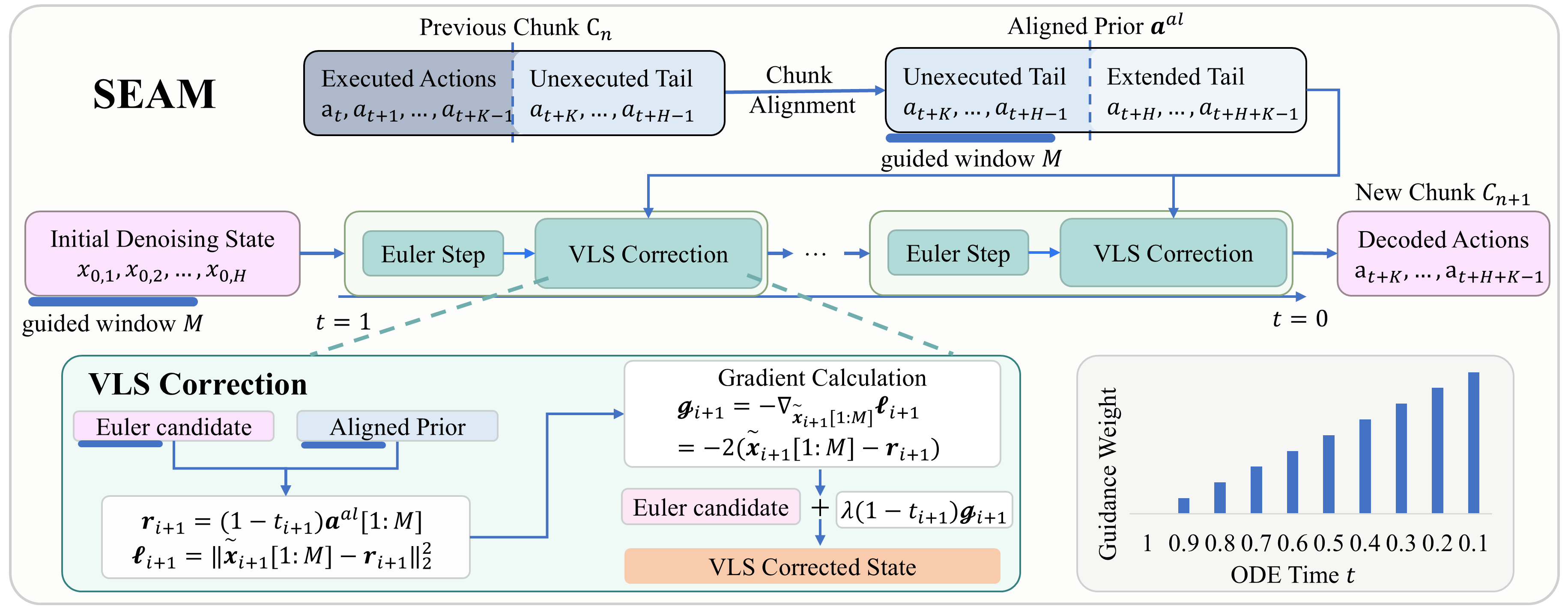}
\caption{Overview of SEAM. The previous decoded chunk $\mathbf{c}_n$ is split into an executed prefix and an unexecuted tail. SEAM extends this tail into the aligned prior $\mathbf{a}^{\text{al}}$ for the next chunk $\mathbf{c}_{n+1}$. Starting from a standard-Gaussian denoising state, the policy runs Euler steps for the new chunk, while VLS evaluates a closed-form loss on the guided window $M$ of each post-Euler candidate and corrects it toward the aligned prior. The correction is scaled by $\lambda(1{-}t_{i+1})$, giving weak guidance at noisy early states and stronger correction near the action manifold, without backpropagating through the policy.}
\label{fig:pipeline}
\end{figure*}

\section{Related Work}

\subsection{Generalist, Chunked, and Generative Robot Policies}
Large robot policies increasingly connect visual observations, language instructions, and low-level actions in a single policy interface, from RT-1 and RT-2 to OpenVLA, Octo, $\pi_0$, $\pi_{0.5}$, and LingBot-VLA~\citep{brohan2022rt1, brohan2023rt2, kim2024openvla, octo2024, black2024pi0, pi05_2025, wu2026lingbotvla}. Many of these systems use \textit{action chunking}, which predicts multiple future controls and executes only a prefix, to obtain locally coherent motion plans~\citep{zhao2023act, chi2023diffusion}. At the same time, recent generative action models use diffusion or flow matching to represent multiple feasible continuations in continuous control, including diffusion VLA policies~\citep{wen2025lladavla, wen2025dvla} and flow-matching action policies~\citep{jia2026a2a, yashima2026hiflow}. This expressiveness is useful for contact-rich tasks, but it also makes independently sampled adjacent chunks vulnerable to choosing incompatible modes, producing the chunk-boundary artifact targeted by SEAM. Rather than scaling the backbone or dataset, SEAM improves cross-chunk consistency at inference time without retraining the base policy or changing the training data.

\subsection{Cross-Chunk Consistency and Inference-Time Steering}
Cross-chunk methods differ in how strongly they constrain the next chunk. Temporal ensembling averages overlaps but cannot select between incompatible modes. RTC~\citep{black2025rtc} uses $\Pi$GDM-style posterior guidance~\citep{song2021score, chung2023dps} and backpropagates through the policy during denoising; BID searches for a consistent continuation by rejection sampling~\citep{liu2024bid}; Legato, ABPolicy, and FASTER require retraining, action-space redesign, or solver-specific timing changes~\citep{liu2026legato, lu2026faster}. More broadly, classifier and classifier-free guidance steer generative sampling at inference time~\citep{dhariwal2021diffusion, ho2022cfg}, while inverse-problem methods guide diffusion trajectories with measurement constraints~\citep{meng2022sdedit, lugmayr2022repaint, kawar2022ddnm, chung2023dps}. SEAM instead targets the structural overlap constraint in synchronous chunked execution and derives analytical, backpropagation-free gradients from the flow matching ODE.


\section{Preliminaries}

\subsection{Flow Matching for Action Generation}
Flow matching~\citep{lipman2023flow, tong2023cfm} formulates generative modeling as learning a velocity field $v_\theta$ that transports samples from a noise distribution to the data distribution via an ODE. Given data $\mathbf{x}_0 \sim p_{\text{data}}$ and noise $\boldsymbol{\epsilon} \sim \mathcal{N}(\mathbf{0}, \mathbf{I})$, the forward interpolation defines:
\begin{equation}
\mathbf{x}_t = (1-t)\,\mathbf{x}_0 + t\,\boldsymbol{\epsilon}, \quad t \in [0, 1]
\label{eq:interpolation}
\end{equation}
with the conditional velocity field $\mathbf{u}_t = \boldsymbol{\epsilon} - \mathbf{x}_0$. The model $v_\theta(\mathbf{x}_t, t)$ is trained to approximate $\mathbf{u}_t$ by minimizing $\|v_\theta(\mathbf{x}_t, t) - \mathbf{u}_t\|^2$. At inference, actions are generated by sampling $\mathbf{z} \sim \mathcal{N}(\mathbf{0}, \mathbf{I})$ and solving the ODE backward from $t{=}1$ to $t{=}0$ using Euler integration:
\begin{equation}
\mathbf{x}_{t+\Delta t} = \mathbf{x}_t + \Delta t \cdot v_\theta(\mathbf{x}_t, t), \quad \Delta t = -\frac{1}{N}
\label{eq:euler}
\end{equation}
where $N$ is the number of integration steps. The final $\mathbf{x}_0$ yields the predicted action sequence.

\subsection{Action Chunking and Overlap}
An action-chunked policy $\pi_\theta$ generates a chunk $\mathbf{c} \in \mathbb{R}^{H \times D}$ of $H$ future actions at each prediction step, where $D$ is the action dimension. Only the first $K$ actions are executed before re-prediction, creating an \textit{overlap region} of length $L = H - K$ between consecutive chunks. After executing $\mathbf{c}_n[1{:}K]$, the next chunk $\mathbf{c}_{n+1}$ should ideally satisfy $\mathbf{c}_{n+1}[1{:}L] \approx \mathbf{c}_n[K{+}1{:}H]$ in the overlap region. In practice, this consistency rarely holds because $\mathbf{c}_{n+1}$ is generated from an independent noise sample $\mathbf{z}_{n+1} \sim \mathcal{N}(\mathbf{0}, \mathbf{I})$.

\paragraph{Multimodal action distributions.}
In contact-rich manipulation, the learned action distribution $p(\mathbf{c} \mid \mathbf{o})$ is inherently multimodal: given the same observation, qualitatively distinct strategies (\eg, grasping from above vs.\ from the side) can all be valid.
Flow matching captures this by mapping different initial noise samples $\mathbf{z}$ to different modes. When consecutive chunks sample $\mathbf{z}_n, \mathbf{z}_{n+1}$ independently, they may converge to incompatible modes even under similar observations, producing a cross-chunk inconsistency that we refer to as \textit{multimodal bifurcation}.


\section{SEAM: Smooth Execution of Action-Chunked Motion}

\subsection{Overview}
SEAM is a plug-and-play inference method for flow matching VLAs. Given the previous chunk's unexecuted tail, it constructs an aligned prior and applies closed-form VLS corrections to the overlap prefix at every Euler step, without retraining the policy, adding auxiliary networks, or backpropagating through the policy at inference time. We use $\mathbf{c}$ for decoded action chunks, $\mathbf{a}$ for executed actions or action-space references, and $\mathbf{x}_i$ for the discrete denoising solver state. Figure~\ref{fig:pipeline} shows the complete pipeline.

\subsection{Aligned Prior Construction}
Given the previous decoded chunk $\mathbf{c}_n$, we define the unexecuted tail as $\mathbf{a}^{\text{tail}}=\mathbf{c}_n[K{+}1{:}H]\in\mathbb{R}^{L \times D}$, where $L = H - K$. We extend this tail to chunk length $H$ by repeating its last entry, producing the \textit{aligned prior}:
\begin{equation}
\mathbf{a}^{\text{al}} = \text{Extend}(\mathbf{a}^{\text{tail}}, H) \in \mathbb{R}^{H \times D}.
\label{eq:aligned_prior}
\end{equation}
The aligned prior is the only piece of cross-chunk state that VLS uses. Although $\mathbf{a}^{\text{al}}$ is represented at chunk length $H$, the main method guides only the first $M \leq L$ overlap actions; repeating the last tail action simply makes the prior length-compatible with the generated chunk. In the main VLS-only setting, the next chunk is still initialised from \textit{standard} Gaussian noise $\mathbf{z} \sim \mathcal{N}(\mathbf{0}, \mathbf{I})^{H \times D}$; the prior acts only as a guidance target during denoising, leaving the policy's marginal sampling process largely intact.

The repeated-tail extension is intentionally conservative. It does not predict a new future trajectory beyond the observed overlap and it does not impose a terminal waypoint on the generated chunk. Instead, it provides a shape-compatible reference whose reliable part is the previous chunk's unexecuted tail. Since VLS applies guidance only to the first $M$ overlap positions, the repeated entries outside this guided prefix mainly serve as a simple padding rule that avoids introducing an additional extrapolator or learned continuation model.

\subsection{Velocity-Guided Loss Steering (VLS)}
VLS operates \textit{during} ODE integration, steering the guided overlap prefix toward the aligned prior. The term ``velocity-guided'' reflects that the correction is inserted into the flow matching trajectory governed by the learned velocity field $v_\theta$, using the linear velocity-path interpolation in Eq.~\ref{eq:interpolation} to define the target at each ODE time. Let $M=\min(L,L_{\max})$ be the number of overlap actions that receive guidance, where $L_{\max}$ is an optional maximum window length. We write the reverse Euler schedule as $t_i=1-i/N$ and let $\mathbf{x}_i$ denote the latent state of the new chunk before the $i$-th Euler step. Thus, $\mathbf{c}_n$ and $\mathbf{a}^{\text{al}}$ are action-space sequences, while $\mathbf{x}_i$ is the current denoising state with the same $H \times D$ chunk shape. If the policy were generating the same trajectory as the aligned prior $\mathbf{a}^{\text{al}}$, then the guided portion of the latent at time $t_i$ should sit near $(1{-}t_i)\,\mathbf{a}^{\text{al}}$. We adopt this as a closed-form, time-varying consistency target:
\begin{equation}
\mathbf{r}_i = (1-t_i) \cdot \mathbf{a}^{\text{al}}[1{:}M].
\label{eq:vls_target}
\end{equation}
This target is a local consistency surrogate, not an exact conditional posterior target: the true reverse state also contains noise and task-conditioned dynamics from the learned velocity field, which VLS deliberately leaves to the policy while applying only a lightweight overlap correction.
At each ODE step, we first compute the standard Euler candidate
\begin{equation}
\widetilde{\mathbf{x}}_{i+1}
= \mathbf{x}_i + (t_{i+1}-t_i)\,v_\theta(\mathbf{x}_i,t_i).
\end{equation}
We then view VLS as minimizing a local quadratic consistency objective on the guided window at the updated ODE time,
\begin{equation}
\ell_{i+1} = \left\|\widetilde{\mathbf{x}}_{i+1}[1{:}M] - \mathbf{r}_{i+1}\right\|_2^2,
\label{eq:vls_loss}
\end{equation}
whose negative gradient with respect to $\widetilde{\mathbf{x}}_{i+1}[1{:}M]$ is available in closed form because $\mathbf{r}_{i+1}$ depends only on the ODE time and the aligned prior, not on an inner optimization or automatic-differentiation pass through the policy network. VLS applies the correction to the guided window only:
\begin{gather}
\mathbf{g}_{i+1} = -\nabla \ell_{i+1}
= -2\bigl(\widetilde{\mathbf{x}}_{i+1}[1{:}M] - \mathbf{r}_{i+1}\bigr), \label{eq:vls_grad} \\
\mathbf{x}_{i+1}[1{:}M] =
\widetilde{\mathbf{x}}_{i+1}[1{:}M]
+ \lambda(1-t_{i+1}) \cdot \mathbf{g}_{i+1}. \label{eq:vls_update}
\end{gather}
Unguided positions keep their Euler-candidate values. The factor $\lambda(1{-}t_{i+1})$ makes the correction small at the noisy beginning of the reverse ODE and larger as the latent approaches the action manifold. This schedule avoids over-constraining early denoising states, while still nudging the final overlap toward the previous chunk's tail. Table~\ref{tab:seam_algorithm} summarizes the complete chunk-level inference procedure.

\begin{table}[h]
\centering
\begingroup
\setlength{\tabcolsep}{2pt}
\small
\begin{tabular}{@{}p{0.97\columnwidth}@{}}
\toprule
\textbf{SEAM inference for one chunk query} \\
\midrule
\textbf{Input:} observation $o$, previous chunk tail $\mathbf{a}^{\text{tail}}$, policy velocity field $v_\theta$, horizon $H$, execution length $K$, guided window $M$, strength $\lambda$, ODE steps $N$. \\
\textbf{1.} Build the aligned prior $\mathbf{a}^{\text{al}}=\text{Extend}(\mathbf{a}^{\text{tail}},H)$ by repeating the last tail action. \\
\textbf{2.} Sample the initial latent $\mathbf{x}_0 \sim \mathcal{N}(\mathbf{0},\mathbf{I})^{H\times D}$ and set $t_i=1-i/N$. \\
\textbf{3.} For each ODE step $i=0,\ldots,N{-}1$: \\
\quad \textbf{3.1} Compute the standard Euler candidate $\widetilde{\mathbf{x}}_{i+1}=\mathbf{x}_i+(t_{i+1}-t_i)v_\theta(\mathbf{x}_i,t_i,o)$. \\
\quad \textbf{3.2} Form the overlap target $\mathbf{r}_{i+1}=(1-t_{i+1})\mathbf{a}^{\text{al}}[1{:}M]$. \\
\quad \textbf{3.3} Apply the closed-form VLS update in Eq.~\ref{eq:vls_update} to $\widetilde{\mathbf{x}}_{i+1}[1{:}M]$ and keep all unguided positions unchanged. \\
\textbf{4.} Decode the final chunk $\mathbf{x}_N$ as the action prediction and execute its first $K$ actions. \\
\bottomrule
\end{tabular}
\endgroup
\caption{Inference procedure for SEAM. The only cross-chunk state is the previous chunk's unexecuted tail, and the guidance correction is computed directly inside the denoising loop without policy-network backpropagation.}
\label{tab:seam_algorithm}
\end{table}

\begin{table*}[t]
\centering
\begingroup
\setlength{\tabcolsep}{1.2mm}
\small
\begin{tabular}{@{}lccccc@{\hspace{4pt}}cccc@{}}
\toprule
\textbf{Method} & \multicolumn{5}{c}{\textbf{Execution quality}} & \multicolumn{4}{c}{\textbf{Denoising-loop cost}} \\
\cmidrule(lr){2-6}\cmidrule(l){7-10}
& \textbf{Succ.\%}$\uparrow$ & \textbf{BJ}$\downarrow$ & \textbf{IJ}$\downarrow$ & \textbf{CD}$\downarrow$ & \textbf{AV$_b$}$\downarrow$ & \textbf{Extra}$\downarrow$ & \textbf{Step}$\downarrow$ & \textbf{Chunk}$\downarrow$ & \textbf{D-Cost}$\downarrow$ \\
\midrule
$\pi_{0.5}$ (Baseline) & 94.8 & 0.195 & 0.094 & 0.172 & 0.165 & 0.000 & 28.225 & 282.2 & $1.00\times$ \\
+ ACT-TE & 82.7 & \textbf{0.031} & \textbf{0.031} & \textbf{0.062} & \textbf{0.006} & 0.000 & 28.225 & 282.2 & $1.00\times$ \\
+ RTC & 95.1 & 0.090 & 0.075 & 0.089 & 0.094 & 6.236 & 34.460 & 344.6 & $1.22\times$ \\
+ \methodname{} & \textbf{95.7} & 0.141 & 0.074 & 0.126 & 0.094 & 0.371 & 28.595 & 286.0 & $1.01\times$ \\
\bottomrule
\end{tabular}
\endgroup
\caption{Cost-aware main results on LIBERO-10. BJ: boundary jerk. IJ: interior jerk. CD: chunk transition discontinuity. AV$_b$: acceleration variance at boundaries. Extra and Step report cached-prefix denoising-loop milliseconds per Euler step; Chunk reports milliseconds per 10-step chunk query; D-Cost is relative to base per-chunk denoising latency. Success is reported to assess task preservation, while lower motion-quality values indicate smoother execution.}
\label{tab:main}
\end{table*}

\paragraph{Design choices and efficiency.} VLS exposes two local design choices: which action dimensions to guide and how much of the overlap to constrain. It can be applied to any subset of action dimensions and over a bounded prefix of the overlap window ($M \leq L$), allowing the method to trade correction footprint against policy freedom. This locality is also what keeps VLS inexpensive. For a guided dimension set of size $D_g$, each ODE step adds only slicing, subtraction, scalar weighting, and an overwrite on an $M \times D_g$ block, giving $\mathcal{O}(N \cdot M \cdot D_g)$ additional scalar operations per chunk query. Because the consistency gradient in Eq.~\ref{eq:vls_grad} is computed in closed form from the post-Euler candidate and the aligned prior, VLS does not require a policy-network backward pass or activation storage for gradient computation. Concrete hyperparameter choices and measured denoising-loop costs are given in the experimental setup.


\section{Experiments}
\label{sec:experiments}

\subsection{Experimental Setup}

\paragraph{Benchmark.} We evaluate on LIBERO-10~\citep{libero2024}, a suite of 10 long-horizon tabletop manipulation tasks in the LIBERO simulation environment. Each task requires multi-step reasoning with diverse objects, involving skills such as picking, placing, opening, and closing.

\paragraph{Backbone Policy.} We use $\pi_{0.5}$~\citep{pi05_2025} as our base VLA policy. $\pi_{0.5}$ employs a PaliGemma-2B~\citep{beyer2024paligemma} vision-language model with a SigLIP~\citep{zhai2023siglip} vision encoder and a Gemma-300M~\citep{team2024gemma} action expert. On LIBERO-10, it generates action chunks of size $H{=}50$ using $N{=}10$ Euler ODE steps, with $K{=}10$ actions executed per chunk (overlap $L{=}40$); the policy is fine-tuned on the LIBERO-10 training set.

\paragraph{SEAM Setting.} Unless otherwise noted, SEAM uses VLS guidance with $\lambda{=}0.1$, linear $(1{-}t)$ decay at every ODE step, guided window $M{=}20$ over the first overlap positions, all physical action dimensions, and standard Gaussian initialization. The aligned prior repeats the final unexecuted-tail action to fill the remaining $H{-}L$ positions.

\paragraph{Baseline settings.} All comparison methods use their official or author-recommended inference settings rather than hyperparameters selected to favor SEAM. ACT temporal ensembling uses the standard coefficient $k{=}0.01$ and queries the policy at every environment step. RTC uses the recommended $\Pi$GDM configuration for action-chunked flow policies, maximum guidance weight $\beta{=}5$, and exponential prefix-attention decay.

\paragraph{Evaluation Protocol.} We report task success rate and four motion quality metrics on executed, post-processed action sequences. All rows use 130 episodes per task. We define the per-step jerk as
\begin{equation}
j_t = \left\|\mathbf{a}_{t+1}-2\mathbf{a}_t+\mathbf{a}_{t-1}\right\|_2
\end{equation}
at step $t$. We use $B=\{t \mid t\bmod K=0,\ t>0\}$ for the set of chunk boundaries and $I$ for the remaining valid centers. Using these sets, the four motion metrics are defined as follows:
\begin{align}
\mathrm{BJ}&=\frac{1}{|B|}\sum_{t\in B} j_t,\\
\mathrm{IJ}&=\frac{1}{|I|}\sum_{t\in I} j_t,\\
\mathrm{CD}&=\frac{1}{|B|}\sum_{t\in B}\left\|\mathbf{a}_t-\mathbf{a}_{t-1}\right\|_2,\\
\mathrm{AV}_b&=\operatorname{Var}\{j_t\mid t\in B\}.
\end{align}
Metrics are accumulated over valid rollout steps within each task and then averaged across the ten tasks.

\paragraph{Timing Protocol.} We report denoising-loop timing on a local RTX 3090 in Table~\ref{tab:main}. The denoising-loop columns measure the cached-prefix ODE loop for one chunk generation with $N{=}10$ Euler steps, isolating the algorithmic overhead added inside the action-generation loop.

\subsection{Main Results}

We compare SEAM against the baseline $\pi_{0.5}$ policy, ACT temporal ensembling, and RTC on LIBERO-10 under matched evaluation settings (Table~\ref{tab:main}).

\begin{figure}[t]
\centering
\includegraphics[width=\columnwidth]{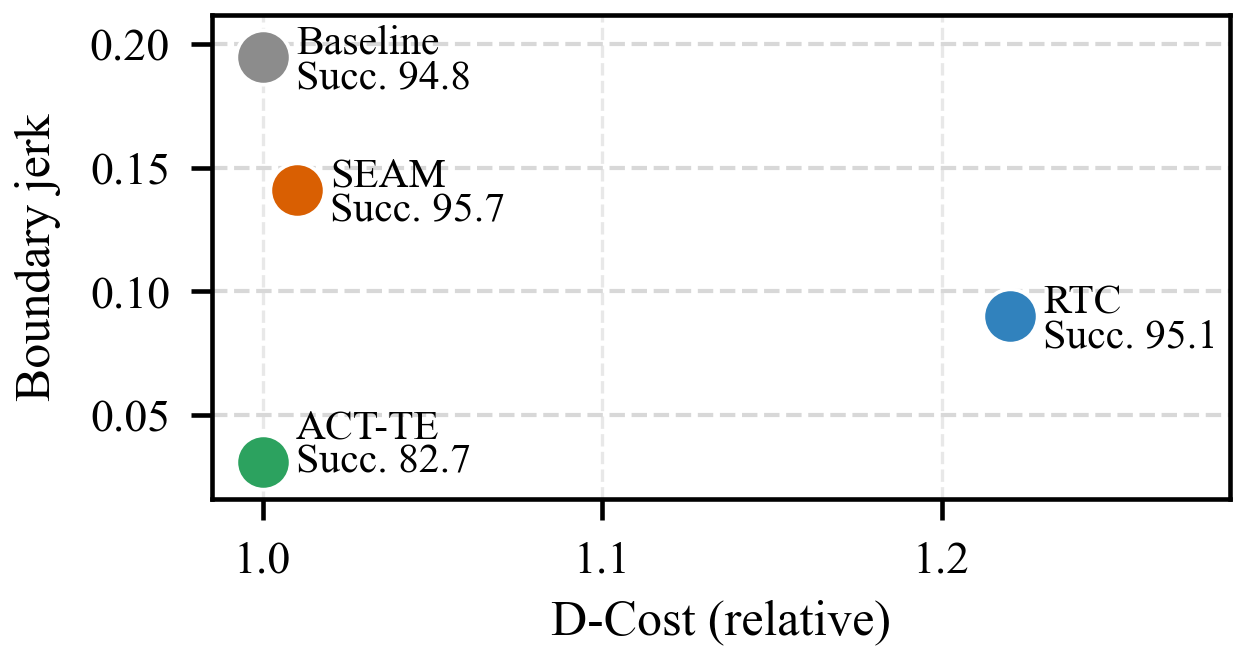}
\caption{Cost--smoothness trade-off. D-Cost denotes relative cached-prefix denoising-loop cost; lower boundary jerk indicates smoother chunk transitions. Point labels report task success rate. SEAM lowers boundary jerk at near-baseline cost, while RTC and ACT-TE trade higher cost or lower success for stronger smoothing.}
\label{fig:operating_point}
\end{figure}

SEAM reduces boundary jerk by 27.7\% (0.195~$\to$~0.141), interior jerk by 21.3\% (0.094~$\to$~0.074), chunk transition discontinuity by 26.7\% (0.172~$\to$~0.126), and AV$_b$ by 43.0\% (0.165~$\to$~0.094), while preserving baseline-level task success (95.7\% vs.\ 94.8\%). The comparison reveals a trade-off among denoising-loop cost, task success, and smoothness across cross-chunk methods. RTC reaches stronger boundary jerk reduction ($-54\%$, to 0.090) and chunk discontinuity reduction ($-48\%$, to 0.089) with 95.1\% success and the same AV$_b$ as SEAM (0.094), but its $\Pi$GDM guidance requires one automatic-differentiation backward pass per ODE step and increases per-chunk denoising latency from 282.2\,ms to 344.6\,ms ($1.22\times$). ACT-TE achieves the most aggressive smoothing (BJ 0.031, $-84\%$, AV$_b$ 0.006) under its every-step query setting, but drops to 82.7\% success. In contrast, SEAM removes the backward pass, adds only 3.8\,ms to the denoising loop per 10-step chunk, and keeps denoising-loop latency at $1.01\times$ the baseline while reducing boundary artifacts and preserving comparable task success. Figure~\ref{fig:operating_point} visualizes this trade-off.

The task-level success rates breakdown in Table~\ref{tab:per_task_success} shows that SEAM preserves aggregate task success without uniformly improving every task: it is strongest on several tasks (T1--T3 and T5), ties the best comparison on T6, and remains below the best comparison method on some tasks (T7--T10). We therefore treat task success, smoothness, and denoising-loop cost jointly rather than as a single-metric ranking. Since T1 exhibits the largest gap among the compared methods, the later qualitative T1 failure-pattern analysis examines representative rollouts to explain how the other methods fail differently on this task.

\begin{table}[t]
\centering
\begingroup
\setlength{\tabcolsep}{3pt}
\small
\begin{tabular}{@{}lrrrr@{}}
\toprule
\textbf{Task} & \textbf{Base.\%} & \textbf{SEAM\%} & \textbf{RTC\%} & \textbf{ACT-TE\%} \\
\midrule
T1 & 91.5 & \textbf{99.2} & 90.8 & 58.5 \\
T2 & 98.5 & \textbf{100.0} & 93.8 & 93.8 \\
T3 & 96.9 & \textbf{100.0} & 96.9 & 93.1 \\
T4 & \textbf{97.7} & 96.9 & 96.2 & 91.5 \\
T5 & 92.3 & \textbf{95.4} & 94.6 & 87.7 \\
T6 & 99.2 & \textbf{100.0} & \textbf{100.0} & 97.7 \\
T7 & 93.8 & 90.8 & \textbf{99.2} & 80.0 \\
T8 & \textbf{97.7} & 96.9 & \textbf{97.7} & 81.5 \\
T9 & 83.8 & 83.8 & \textbf{86.2} & 55.4 \\
T10 & \textbf{96.2} & 93.8 & 95.4 & 87.7 \\
\midrule
\textbf{Overall} & 94.8 & \textbf{95.7} & 95.1 & 82.7 \\
\bottomrule
\end{tabular}
\endgroup
\caption{Per-task success rates on LIBERO-10. T1--T10 denote the fixed LIBERO-10 task order used for the aggregate results in Table~\ref{tab:main}. Best values in each row are in boldface; ties are bolded for all tied methods.}
\label{tab:per_task_success}
\end{table}

\begin{figure*}[t]
\centering
\includegraphics[width=\textwidth]{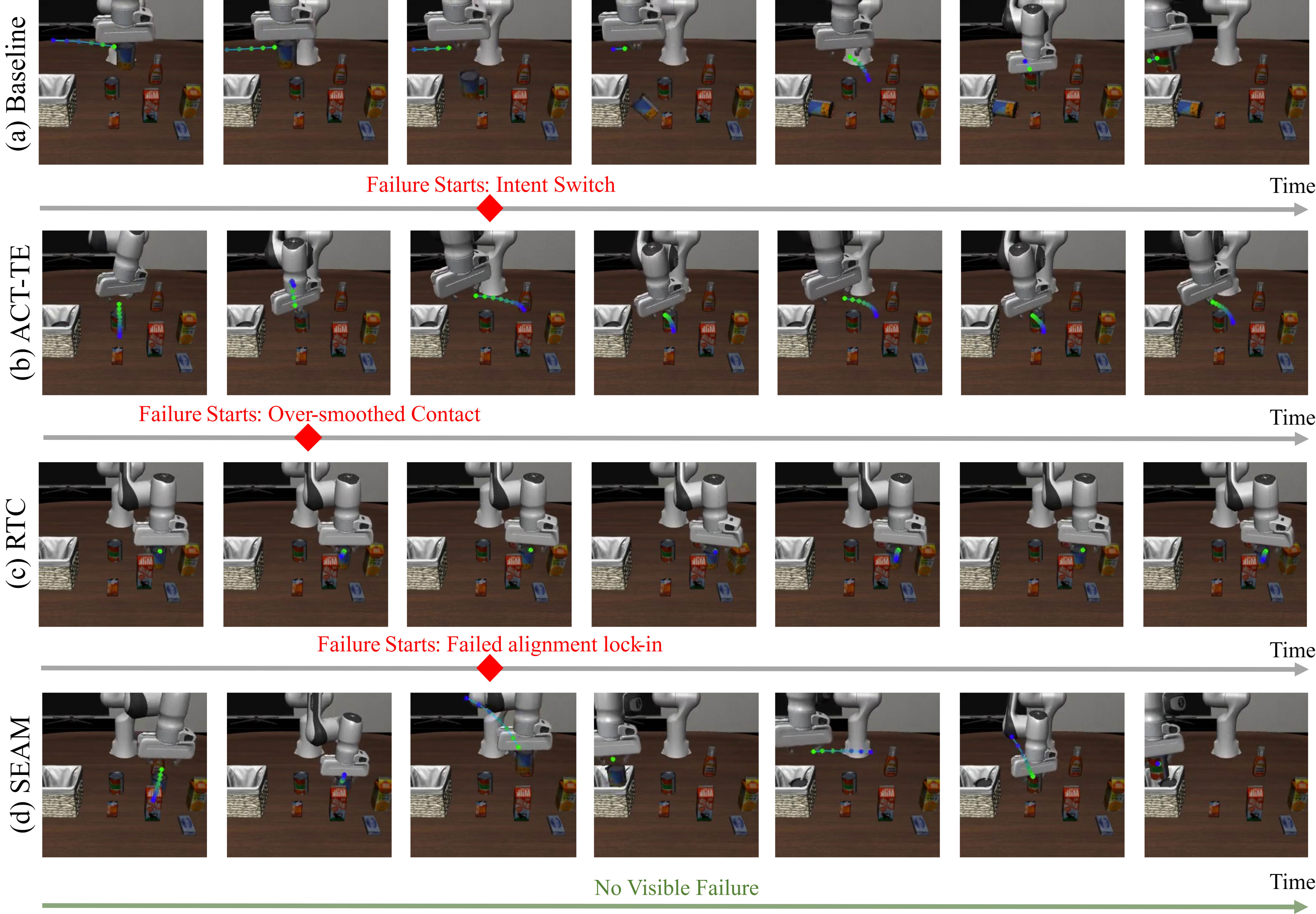}
\caption{Qualitative T1 failure analysis. Representative T1 rollouts explain the largest gap in Table~\ref{tab:per_task_success}: the baseline shows intent inconsistency, ACT-TE over-smooths contact timing, and RTC can lock into a failed alignment. SEAM preserves corrective action freedom while improving cross-chunk consistency.}
\label{fig:pattern_analysis}
\end{figure*}

\subsection{Ablation Study}

\begin{table}[t]
\centering
\begingroup
\setlength{\tabcolsep}{1mm}
\small
\begin{tabular}{llccccc}
\toprule
\textbf{Dims} & \textbf{$\lambda$} & \textbf{$M$} & \textbf{Succ.\%}$\uparrow$ & \textbf{BJ}$\downarrow$ & \textbf{IJ}$\downarrow$ & \textbf{CD}$\downarrow$ \\
\midrule
none & -- & -- & 94.8 & 0.195 & 0.094 & 0.172 \\
\midrule
all & 0.05 & 20 & 94.8 & 0.162 & 0.082 & 0.145 \\
all & 0.10 & 20 & \textbf{95.7} & 0.141 & 0.074 & \textbf{0.126} \\
all & 0.15 & 20 & 92.8 & \textbf{0.140} & 0.071 & 0.127 \\
all & 0.20 & 20 & 89.5 & 0.147 & \textbf{0.067} & 0.135 \\
\midrule
pos & 0.05 & 20 & 94.3 & 0.179 & 0.089 & 0.159 \\
pos & 0.10 & 20 & 94.0 & 0.162 & 0.085 & 0.143 \\
pos & 0.15 & 20 & 92.6 & 0.166 & 0.085 & 0.145 \\
pos & 0.20 & 20 & 88.2 & 0.168 & 0.083 & 0.152 \\
\bottomrule
\end{tabular}
\endgroup
\caption{Ablation of SEAM guidance strength and dimension selection on LIBERO-10. ``pos'': guidance on position dimensions only. ``all'': all action dimensions. All guided rows use standard Gaussian initialization with $M{=}20$. Best guided values are in boldface.}
\label{tab:ablation}
\end{table}

\paragraph{VLS strength trade-off.} We first sweep the guidance strength $\lambda$ at fixed $M{=}20$ for both all-dimension and position-only guidance, with the results shown in Table~\ref{tab:ablation}. In the all-dimension rows, increasing $\lambda$ generally strengthens smoothing but quickly erodes task reliability. $\lambda{=}0.05$ reaches 94.8\% success with moderate smoothing, while the main $\lambda{=}0.1$ setting gives the best balance: 95.7\% success with BJ 0.141 and CD 0.126. Larger strengths only marginally improve some jerk metrics, if at all, but reduce success to 92.8\% at $\lambda{=}0.15$ and 89.5\% at $\lambda{=}0.2$. This supports using a weak closed-form correction rather than forcing the denoising trajectory toward the previous chunk too aggressively.

\paragraph{Dimension selection matters.} We next test whether VLS should guide all action dimensions or only position dimensions, with the matched sweeps shown in Table~\ref{tab:ablation}. Restricting guidance to position dimensions (``pos-only'') is more conservative but less effective at smoothing than the matched all-dimension setting. At $\lambda{=}0.1$ and $M{=}20$, pos-only guidance yields BJ 0.162 and CD 0.143 with 94.0\% success, whereas all-dimension guidance yields BJ 0.141 and CD 0.126 while preserving task success at 95.7\%, supporting the bounded-window all-dimension setting used in the main results.

\begin{table}[t]
\centering
\begingroup
\setlength{\tabcolsep}{1mm}
\small
\begin{tabular}{lcccc}
\toprule
\textbf{Setting} & \textbf{$M$} & \textbf{Succ.\%}$\uparrow$ & \textbf{BJ}$\downarrow$ & \textbf{CD}$\downarrow$ \\
\midrule
Baseline & -- & 94.8 & 0.195 & 0.172 \\
SEAM & 2  & 95.5 & 0.148 & 0.142 \\
SEAM & 4  & 94.8 & 0.156 & 0.146 \\
SEAM & 6  & 96.3 & 0.147 & 0.137 \\
SEAM & 8  & 95.0 & 0.149 & 0.138 \\
SEAM & 10 & 95.9 & 0.148 & 0.135 \\
SEAM & 12 & 95.1 & 0.153 & 0.140 \\
SEAM & 14 & 95.7 & 0.154 & 0.141 \\
SEAM & 16 & 94.7 & 0.147 & 0.134 \\
SEAM & 18 & 95.5 & 0.148 & 0.135 \\
SEAM & 20 & 95.7 & \textbf{0.141} & \textbf{0.126} \\
\bottomrule
\end{tabular}
\endgroup
\caption{Full-scale sweep of SEAM guided-window length $M$ on LIBERO-10. All SEAM rows use all action dimensions with $\lambda{=}0.1$ and standard Gaussian initialization. Best values are in boldface.}
\label{tab:window_sweep}
\end{table}

\paragraph{Window length controls boundary smoothing.} The Table~\ref{tab:window_sweep} shows that every tested $M$ reduces BJ and CD relative to the baseline, while task success remains broadly stable from 94.7\% to 96.3\%. Larger windows tend to reduce CD more strongly, and $M{=}20$ gives the strongest smoothing (BJ 0.141, CD 0.126). This supports treating $M$ primarily as a smoothness knob: increasing the guided window places more of the overlap under the consistency target, while task success remains stable within the tested range.

\paragraph{Qualitative T1 failure patterns.}
To explain why T1 shows the largest per-task gap in Table~\ref{tab:per_task_success}, Figure~\ref{fig:pattern_analysis} compares representative T1 rollouts. The unguided baseline can switch intent and drop the object during transfer; ACT-TE can over-smooth contact timing and delay accurate grasping; RTC can preserve a failed early grasp through its stronger continuation constraint. SEAM avoids these failure modes by using weak overlap consistency rather than hard continuation, preserving corrective freedom while improving cross-chunk consistency. This pattern is consistent with the T1 success rates in Table~\ref{tab:per_task_success}.


\section{Conclusion}

We presented SEAM, a training-free inference-time method for improving chunk-boundary consistency in flow matching-based VLA policies. SEAM uses the previous chunk's unexecuted tail as a consistency reference and applies VLS corrections inside the ODE solver without backpropagating through the policy network. On LIBERO-10, SEAM reduces boundary artifacts while preserving task success and keeping denoising-loop cost near the unguided baseline. Overall, SEAM offers a lightweight way to improve cross-chunk consistency in action-chunked VLA policies, balancing smoother transitions with task success and denoising-loop cost.

\section*{Acknowledgments}
This work was supported by Guangdong Provincial Natural Science Foundation for Outstanding Youth Team Project (No. 2024B1515040010), National Natural Science Foundation of China under Grant U23A20391, 62372188, and Guangdong Natural Science Foundation under Grant 2024A1515010100.

\bibliography{references}

\end{document}